\pdfoutput=1
\documentclass[conference]{IEEEtran}
\IEEEoverridecommandlockouts
\usepackage{cite}
\usepackage{amsmath,amssymb,amsfonts}
\usepackage{graphicx}
\usepackage{textcomp}\usepackage{pifont}
\usepackage{xcolor}
\usepackage{booktabs,tabularx,array,multirow}
\usepackage{microtype}
\usepackage{tikz}\usetikzlibrary{positioning,shapes.geometric,arrows.meta,fit,backgrounds,calc}
\definecolor{cSpec}{HTML}{2A78D6}\definecolor{cDec}{HTML}{EB6834}\definecolor{cInk}{HTML}{26313D}
\definecolor{cFail}{HTML}{C0392B}\definecolor{cPass}{HTML}{2E8B57}\definecolor{cGold}{HTML}{E0A100}
\definecolor{gOne}{HTML}{4A5A70}\definecolor{gTwo}{HTML}{2C7A62}\definecolor{gThree}{HTML}{6B4E9B}
\usepackage[scaled=.92]{helvet}
\usepackage[hidelinks]{hyperref}
\newcommand{\tabnote}[1]{\par\vspace{3pt}\begin{minipage}{\linewidth}\scriptsize\raggedright\textit{Note:} #1\end{minipage}}
\newcommand{\Qop}{Q_{\mathrm{all}}}

\begin{document}

\title{Can System-1-Style Decision Models Replace Dedicated Rerankers in RAG? A Systems Evaluation}

\author{\IEEEauthorblockN{Qiaomu Li, Ying Xie}
\IEEEauthorblockA{\textit{College of Computing and Software Engineering, Kennesaw State University, Marietta, GA, USA}\\
qli12@students.kennesaw.edu, yxie2@kennesaw.edu}}

\maketitle

\begin{abstract}
Retrieval-augmented generation (RAG) systems often use a dedicated reranker to choose which retrieved candidates reach the language model. Reranking improves relevance but adds latency, so teams look for faster alternatives. System-1-style decision models make direct, structured predictions instead of generating long responses. Could one provide comparable reranking quality with less inference overhead? We first screen 21 open-weight models on 11 benchmarks, then compare a promising decision model with a dedicated reranker on identical candidate lists. \textbf{The default decision-model setup fails on many long candidate sets, so scoring only the requests it completes gives a misleading ranking.} Limiting candidate text length largely restores reliability and makes some configurations competitive. A frozen evaluation on new retrieval data supports the reliability and ranking results, but confirms the latency requirement for only one of the two selected configurations. Across workloads, which model ranks better changes, and a tuned reranker remains faster and uses far less GPU memory. \textbf{A decision model can perform competitive reranking, but replacing a reranker requires weighing quality, latency and memory together, and teams should tune the reranker they already run first.}
\end{abstract}


\section{Introduction}
Retrieval-augmented generation (RAG) systems often retrieve more candidates than they can pass to the downstream language model. A dedicated reranker helps select the most relevant candidates for the final context~\cite{robertson2009bm25,lewis2020rag,gao2023ragsurvey,nogueira2019bert}. This usually improves the quality of the retrieved context, but processing a large candidate set can add substantial latency. In practice, teams must balance relevance, response time and GPU resources.

System-1-style decision models offer a possible alternative. Instead of generating a free-form answer, they return scores or selections over predefined options. Reranking can also be framed as choosing the best candidates from a set, so these models may be able to do the same job with less inference overhead. Yet specialized rerankers are built for relevance ranking, and a decision model must still process the query and every candidate text. A lower latency is therefore a hypothesis, not a given.

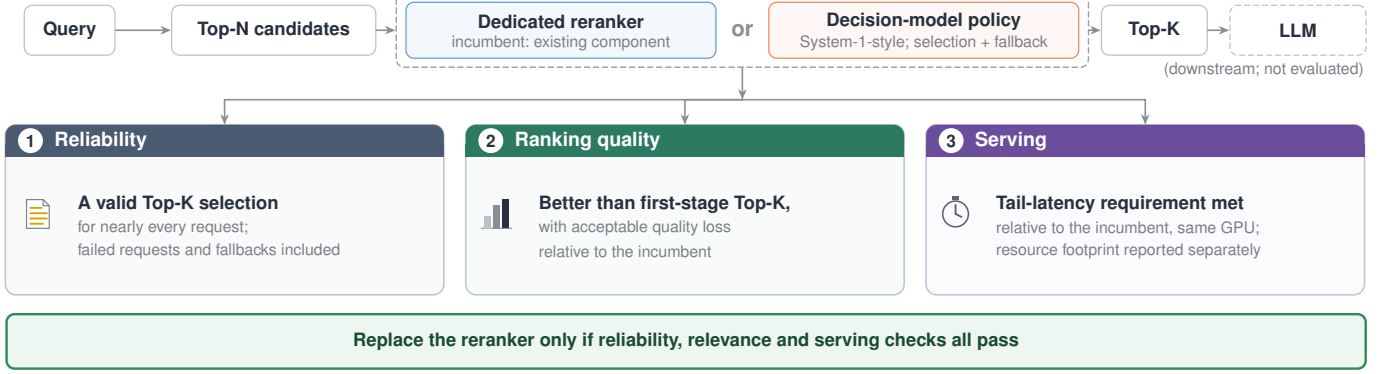
\begin{figure*}[t]
\centering\tikzset{
  pics/doc/.style={code={
    \fill[white, draw=cInk!60, line width=.45pt] (0,0) -- (0.30,0) -- (0.30,0.30) -- (0.22,0.38) -- (0,0.38) -- cycle;
    \draw[cInk!60, line width=.45pt] (0.22,0.38) -- (0.22,0.30) -- (0.30,0.30);
    \foreach \y/\w in {0.27/0.15,0.20/0.22,0.13/0.22,0.06/0.17} \draw[#1, line width=.7pt] (0.05,\y) -- ++(\w,0);
  }},
  pics/doc/.default=cInk!35,
  pics/bars/.style={code={
    \draw[cInk!55, line width=.45pt] (0,0) -- (0.40,0);
    \foreach \x/\h/\c in {0.04/0.16/cInk!30,0.16/0.30/cInk!55,0.28/0.38/cInk} \fill[\c] (\x,0) rectangle ++(0.08,\h);
  }},
  pics/watch/.style={code={
    \draw[cInk!75, line width=.6pt] (0.19,0.19) circle (0.17);
    \fill[cInk!75] (0.15,0.38) rectangle (0.23,0.43);
    \draw[cInk!75, line width=.7pt] (0.19,0.19) -- (0.19,0.31) (0.19,0.19) -- (0.27,0.15);
  }},
}
\begin{tikzpicture}[font=\sffamily\fontsize{7.5}{9}\selectfont, >={Stealth[length=4pt,width=3.2pt]}, line cap=round,
  every node/.style={inner sep=0pt, text=cInk},
  box/.style={draw=cInk!28, line width=.5pt, rounded corners=3pt, fill=white, minimum height=9mm},
  sbox/.style={box, draw=cSpec!70, fill=cSpec!5},
  dbox/.style={box, draw=cDec!75, fill=cDec!5},
  t/.style={font=\sffamily\bfseries\fontsize{7.5}{9}\selectfont, anchor=west, text=cInk},
  st/.style={font=\sffamily\fontsize{6.5}{8}\selectfont, anchor=west, text=cInk!62},
  sm/.style={font=\sffamily\fontsize{6.5}{8}\selectfont},
  arr/.style={->, line width=.6pt, draw=cInk!55},
  panel/.style={draw=#1!35, line width=.6pt, rounded corners=4pt, fill=#1!3},
  strip/.style={fill=#1, rounded corners=4pt},
]
\node[box, minimum height=6.6mm, minimum width=12mm] (pq) at (0.85,0) {\textbf{Query}};
\node[box, minimum height=6.6mm, minimum width=27mm] (pr) at (3.55,0) {\textbf{Top-N candidates}};
\node[sbox, minimum height=7.4mm, minimum width=41mm] (ps) at (7.35,0.0) {};
\node[font=\sffamily\bfseries\fontsize{7.5}{9}\selectfont] at ($(ps.center)+(0,0.12)$) {Dedicated reranker};
\node[sm, text=cInk!62] at ($(ps.center)+(0,-0.17)$) {incumbent: existing component};
\node[dbox, minimum height=7.4mm, minimum width=41mm] (pd) at (12.15,0.0) {};
\node[font=\sffamily\bfseries\fontsize{7.5}{9}\selectfont] at ($(pd.center)+(0,0.12)$) {Decision-model policy};
\node[sm, text=cInk!62] at ($(pd.center)+(0,-0.17)$) {System-1-style; selection + fallback};
\node[font=\sffamily\bfseries\fontsize{9}{9}\selectfont, text=cInk!55] at (9.75,0) {or};
\node[box, minimum height=6.6mm, minimum width=14mm] (p5) at (15.2,0) {\textbf{Top-K}};
\node[box, dash pattern=on 2pt off 1.4pt, minimum height=6.6mm, minimum width=18mm] (pg) at (17.1,0) {\textbf{LLM}};
\node[sm, text=cInk!62, anchor=east] at (17.98,-0.52) {(downstream; not evaluated)};
\node[draw=cInk!45, line width=.55pt, dash pattern=on 2pt off 1.4pt, rounded corners=3pt, fit=(ps)(pd), inner xsep=1.2mm, inner ysep=1.1mm] (slot) {};
\draw[arr] (pq) -- (pr); \draw[arr] (pr) -- (slot.west); \draw[arr] (slot.east) -- (p5); \draw[arr] (p5) -- (pg);
\node[sm, text=cInk!55, anchor=west] at (0.05,0.62) {\textbf{Reranker replacement with a fixed retrieval stage.} Same query, same retrieved candidates; only the reranking policy (input handling, model, fallback) changes.};

\foreach \xa/\xb/\c/\n/\h in {0/5.8/gOne/1/Reliability, 6.09/11.89/gTwo/2/Ranking quality, 12.18/17.98/gThree/3/Serving} {
  \begin{scope}[on background layer]
    \draw[panel=\c] (\xa,-3.50) rectangle (\xb,-1.25);
    \fill[strip=\c] (\xa,-1.67) rectangle (\xb,-1.25);
    \fill[\c] (\xa,-1.67) rectangle (\xb,-1.51);
  \end{scope}
  \node[circle, fill=white, minimum size=3.3mm, font=\sffamily\bfseries\fontsize{7}{8}\selectfont, anchor=west] (c\n) at (\xa+0.16,-1.46) {\n};
  \node[font=\sffamily\bfseries\fontsize{8}{9}\selectfont, text=white, anchor=west] at ($(c\n.east)+(1.4mm,0)$) {\h};
}
\coordinate (m) at ($(slot.south)+(0,-0.42)$);
\draw[arr] (slot.south) -- (m);
\draw[arr] (m) -| (2.9,-1.25);
\draw[arr] (m) -| (8.99,-1.25);
\draw[arr] (m) -| (15.08,-1.25);

\pic at (0.28,-2.62) {doc=cGold};
\node[t] at (0.95,-2.30) {A valid Top-K selection};
\node[st] at (0.95,-2.62) {for nearly every request;};
\node[st] at (0.95,-2.92) {failed requests and fallbacks included};
\pic at (6.30,-2.62) {bars};
\node[t] at (7.05,-2.30) {Better than first-stage Top-K,};
\node[st] at (7.05,-2.62) {with acceptable quality loss};
\node[st] at (7.05,-2.92) {relative to the incumbent};
\pic at (12.38,-2.62) {watch};
\node[t] at (13.10,-2.30) {Tail-latency requirement met};
\node[st] at (13.10,-2.62) {relative to the incumbent, same GPU;};
\node[st] at (13.10,-2.92) {resource footprint reported separately};

\node[box, draw=cPass, line width=.9pt, fill=cPass!8, minimum width=17.98cm, minimum height=7.2mm, align=center,
      font=\sffamily\bfseries\fontsize{7.5}{9}\selectfont, text=cPass!55!black, anchor=north west] at (0,-3.74) {Replace the reranker only if reliability, relevance and serving checks all pass};
\end{tikzpicture}
\caption{\textbf{Evaluating a reranker replacement in a fixed RAG pipeline.} For a given query, both the incumbent dedicated reranker and a proposed decision-model policy receive the same retrieved candidate set. We compare the Top-K candidates they return, including failures and fallback behavior, and measure the serving latency of each policy. The retriever is unchanged, which isolates the reranking step. The downstream LLM is shown for context but is not part of the controlled ranking evaluation. A replacement is justified only when reliability, ranking quality and serving requirements are met together.}
\label{fig:pipeline}
\end{figure*}

This leads to a practical question: \textbf{Can a decision model replace a dedicated reranker while meeting the same reliability, quality and latency requirements?} A model may rank well on the requests it completes yet fail on long inputs. It may also improve ranking quality by making extra inference calls, leaving the reranking stage slower or more memory hungry. Ranking scores alone can hide these trade-offs.

We study this question at the reranking stage of the fixed RAG pipeline in Fig.~\ref{fig:pipeline}. Every policy receives the same query and retrieved candidates, so only its input handling, ranking model and failure behavior differ. We first screen a broad set of decision models and dedicated rerankers, then compare a promising pair in a controlled study of input limits, candidate order, multi-stage selection and serving latency. Finally, we test selected configurations on a new retrieval benchmark without retuning them. \textbf{This design separates a model's ability to rerank from its suitability as a production replacement.}

\section{Related Work}
\textbf{Dedicated rerankers.} Many retrieval systems use two stages. A fast retriever first returns a set of candidates, and a dedicated reranker then scores their relevance to the query before the top results move downstream. Rerankers may score query--candidate pairs separately, as cross-encoders do~\cite{nogueira2019bert,nogueira2020monot5}, or rank several candidates jointly, as listwise models do~\cite{ma2023lrl}; their architectures and inference costs differ. These models often improve ranking quality, but their runtime grows with the number and length of the inputs. More efficient designs have reduced that overhead~\cite{nasika2026jina}, which makes a tuned specialist the natural baseline for any proposed replacement.

\textbf{LLMs as rerankers.} General-purpose language models can also rank by prompting. Pointwise methods judge candidates one at a time, pairwise and setwise methods compare small groups, and listwise methods produce an ordering of the whole set~\cite{sun2023rankgpt,ma2023lrl,qin2024prp,zhuang2024setwise}; some approaches further fine-tune the model for ranking~\cite{pradeep2023rankvicuna,pradeep2023rankzephyr}. This offers flexibility without training a reranker for every setting, but inference can be expensive, and results may depend on candidate order~\cite{tang2024found,zheng2024mcq,pezeshkpour2024order,lu2022fantastically} and input length~\cite{liu2024lost,shi2023distracted}. Serving cost also grows with input length~\cite{dao2022flash,kwon2023vllm}.

\textbf{System-1-style decision models.} Recent decision models return bounded outputs, such as a score or a selection among supplied options, rather than long free-form responses. A compact, predictable output makes them appealing wherever a system needs a quick decision, and reranking can be viewed as one such decision. The label borrows from dual-process accounts of fast, intuitive judgment~\cite{stanovich2000individual} and was introduced for this model family by TypeSafe AI; we use it only as an interface-level grouping, since architectures and serving costs vary and we did not audit them. Early preprints explore decision-based reranking for recommendation~\cite{lyu2026decision} and multimodal retrieval~\cite{liu2026multimodal}. A structured interface alone does not establish an advantage over a ranking specialist, because the model must still read every candidate.

\textbf{Research gap.} Reranking studies usually emphasize ranking quality (nDCG~\cite{jarvelin2002ndcg}) on benchmarks such as BEIR~\cite{thakur2021beir} and BRIGHT~\cite{su2025bright}. The open question is whether a decision model can replace a dedicated reranker in RAG while meeting real serving requirements. Long candidate sets, invalid outputs, candidate-order effects and latency can all change which approach is preferable. We therefore evaluate the complete reranking policy with retrieval held fixed, measuring reliability, ranking quality and serving performance together.

\section{Methodology}
\subsection{Model screen and benchmark coverage}\label{sec:screen}
We screened 21 open-weight checkpoints from 0.02B to 35B parameters (nine dedicated rerankers and 12 System-1-style decision models): a diverse sample of locally runnable models with documented scoring or structured-choice interfaces that fit one 80\,GB A100, not an exhaustive list. We did not tune model weights on these labels. Rerankers score each candidate and we sort. Most decision models return a typed choice or pointer over the options shown; Intern-Decision-4B returns a joint binary verdict, Clef-Flash ran jointly and Laya pointwise. Decision models receive the query and candidate texts in BM25 order through their official templates. PPLX, run through its official decision interface (noncausal attention, calibrated probabilities), returns one score per candidate in a single call and we keep the five highest (ties by original order), so single-call PPLX is one forward pass and two-stage PPLX three; other checkpoints return typed choices or pointers. An error or malformed output is a failed request. Table~\ref{tab:bench} lists the 11 benchmarks (10,079 queries).

We restrict the study to locally runnable open-weight checkpoints so that versions, input handling, memory and latency compare on the same hardware. Jev is a hosted API whose provider-side latency we cannot isolate.

\begin{table*}[t]
\centering
\caption{How broad is the screen? Quality of 21 open-weight models on 11 benchmarks, from earlier project stages and not one shared protocol ($\times$100)}\label{tab:overview}
\scriptsize\setlength{\tabcolsep}{3.4pt}
\begin{tabular}{@{}l r rrrr rrrr rrrr@{}}\toprule
 & & \multicolumn{4}{c}{Reasoning and science} & \multicolumn{4}{c}{Multi-hop QA} & \multicolumn{4}{c}{Grounded and domain QA}\\\cmidrule(lr){3-6}\cmidrule(lr){7-10}\cmidrule(l){11-14}
 & Size & BRIGHT & +trunc. & SciFact & QASPER & HotpotQA & MuSiQue & 2Wiki & MH-RAG & GaRAGe & Enterpr. & FinBench & CRAG\\
Model & (B) & \multicolumn{3}{c}{nDCG@5} & FS@5 & \multicolumn{4}{c}{FS@5} & Rel@5 & FS@5 & Hit@5 & F1\\
\midrule
First-stage order, Top-5 & -- & 13.3 & \textcolor{black!30}{--} & 75.0 & 14.6 & 49.8 & 9.9 & 24.4 & 38.9 & 72.9 & 55.8 & 29.4 & 24.8\\
Random Top-5 & -- & 5.8 & \textcolor{black!30}{--} & \textcolor{black!30}{--} & \textcolor{black!30}{--} & \textcolor{black!30}{--} & \textcolor{black!30}{--} & \textcolor{black!30}{--} & \textcolor{black!30}{--} & 72.1 & \textcolor{black!30}{--} & \textcolor{black!30}{--} & \textcolor{black!30}{--}\\
\addlinespace[2pt]\multicolumn{14}{@{}l}{\textit{Dedicated rerankers (9)}}\\[1pt]
\quad MiniLM-L6~\cite{wang2020minilm} & 0.02 & \textcolor{black!30}{--} & \textcolor{black!30}{--} & \textcolor{black!30}{--} & \textcolor{black!30}{--} & 59.5 & 14.3 & 32.6 & \textcolor{black!30}{--} & \textcolor{black!30}{--} & 54.4 & \textcolor{black!30}{--} & \textcolor{black!30}{--}\\
\quad GTE-ModernBERT~\cite{zhang2024mgte} & 0.15 & \textcolor{black!30}{--} & \textcolor{black!30}{--} & \textcolor{black!30}{--} & \textcolor{black!30}{--} & 68.4 & 15.1 & 31.9 & \textcolor{black!30}{--} & \textcolor{black!30}{--} & 56.7 & \textcolor{black!30}{--} & \textcolor{black!30}{--}\\
\quad BGE-v2-m3~\cite{chen2024bgem3} & 0.57 & \textcolor{black!30}{--} & \textcolor{black!30}{--} & 74.7 & \textcolor{black!30}{--} & 68.8 & 16.8 & 34.0 & 35.1 & 84.4 & 53.7 & 50.5 & \textcolor{black!30}{--}\\
\quad Qwen3-Reranker-0.6B~\cite{zhang2025qwen3emb} & 0.6 & 16.8 & \textcolor{black!30}{--} & 76.5 & \textcolor{black!30}{--} & 66.4 & 16.0 & 33.1 & \textcolor{black!30}{--} & \textbf{85.0} & 52.6 & \textcolor{black!30}{--} & \textcolor{black!30}{--}\\
\quad Jina v3.5~\cite{nasika2026jina} & 0.6 & \textbf{18.4} & \textcolor{black!30}{--} & 79.3 & 19.9 & \textbf{72.8} & \textbf{20.0} & 38.5 & \textbf{46.0} & 84.4 & 59.4 & \textbf{59.6} & 27.7\\
\quad Mixedbread v2 & 1.5 & 15.8 & \textcolor{black!30}{--} & \textcolor{black!30}{--} & \textcolor{black!30}{--} & \textcolor{black!30}{--} & \textcolor{black!30}{--} & \textcolor{black!30}{--} & \textcolor{black!30}{--} & 84.9 & \textcolor{black!30}{--} & \textcolor{black!30}{--} & \textcolor{black!30}{--}\\
\quad Qwen3-VL-Reranker~\cite{li2026qwen3vlemb} & 2.1 & \textcolor{black!30}{--} & \textcolor{black!30}{--} & \textcolor{black!30}{--} & \textcolor{black!30}{--} & 69.1 & 18.1 & 34.8 & \textcolor{black!30}{--} & \textcolor{black!30}{--} & 55.1 & \textcolor{black!30}{--} & \textcolor{black!30}{--}\\
\quad Qwen3-Reranker-4B~\cite{zhang2025qwen3emb} & 4.0 & \textcolor{black!30}{--} & \textcolor{black!30}{--} & 79.0 & \textcolor{black!30}{--} & 70.0 & 17.4 & 35.2 & \textcolor{black!30}{--} & 84.2 & 46.3 & \textcolor{black!30}{--} & \textcolor{black!30}{--}\\
\quad zerank-2~\cite{pipitone2025zelo} & 4.0 & \textcolor{black!30}{--} & \textcolor{black!30}{--} & \textcolor{black!30}{--} & \textcolor{black!30}{--} & 63.5 & 14.5 & 33.5 & 41.9 & \textcolor{black!30}{--} & 64.3 & 58.7 & \textcolor{black!30}{--}\\
\addlinespace[2pt]\multicolumn{14}{@{}l}{\textit{System-1-style decision models (12)}}\\[1pt]
\quad Laya & 0.42 & \textcolor{black!30}{--} & \textcolor{black!30}{--} & \textcolor{black!30}{--} & \textcolor{black!30}{--} & 10.8 & 1.2 & 4.1 & \textcolor{black!30}{--} & \textcolor{black!30}{--} & 27.1 & \textcolor{black!30}{--} & \textcolor{black!30}{--}\\
\quad Amos 1 Small & 2 & \textcolor{black!30}{--} & \textcolor{black!30}{--} & 53.3\rlap{$^\dagger$} & \textcolor{black!30}{--} & \textcolor{black!30}{--} & \textcolor{black!30}{--} & \textcolor{black!30}{--} & \textcolor{black!30}{--} & 81.5 & \textcolor{black!30}{--} & \textcolor{black!30}{--} & \textcolor{black!30}{--}\\
\quad Liquid d1-3B & 3.1 & 14.8 & \textcolor{black!30}{--} & 75.8 & \textcolor{black!30}{--} & 65.1 & 15.4 & 39.2 & 30.0 & 83.4 & 51.2 & 53.2 & \textcolor{black!30}{--}\\
\quad Intern-Decision-4B & 4.5 & \textcolor{black!30}{--} & \textcolor{black!30}{--} & \textcolor{black!30}{--} & \textcolor{black!30}{--} & 57.5 & 13.7 & 32.9 & \textcolor{black!30}{--} & \textcolor{black!30}{--} & 0.0\rlap{$^\dagger$} & \textcolor{black!30}{--} & \textcolor{black!30}{--}\\
\quad StartLux-Decision-4B & 4.7 & \textcolor{black!30}{--} & \textcolor{black!30}{--} & 80.8 & \textcolor{black!30}{--} & \textcolor{black!30}{--} & \textcolor{black!30}{--} & \textcolor{black!30}{--} & \textcolor{black!30}{--} & 83.8 & \textcolor{black!30}{--} & \textcolor{black!30}{--} & \textcolor{black!30}{--}\\
\quad Drex-DLM & 8.2 & 7.1\rlap{$^\dagger$} & 11.9 & \textcolor{black!30}{--} & \textcolor{black!30}{--} & \textcolor{black!30}{--} & \textcolor{black!30}{--} & \textcolor{black!30}{--} & \textcolor{black!30}{--} & 83.0 & \textcolor{black!30}{--} & \textcolor{black!30}{--} & \textcolor{black!30}{--}\\
\quad Kev-9B & 9 & 6.6\rlap{$^\dagger$} & 12.1 & \textcolor{black!30}{--} & \textcolor{black!30}{--} & \textcolor{black!30}{--} & \textcolor{black!30}{--} & \textcolor{black!30}{--} & \textcolor{black!30}{--} & 84.7 & \textcolor{black!30}{--} & \textcolor{black!30}{--} & \textcolor{black!30}{--}\\
\quad Bespoke Nimble-9B-v2 & 9 & \textcolor{black!30}{--} & \textcolor{black!30}{--} & \textcolor{black!30}{--} & \textcolor{black!30}{--} & 66.9 & 16.0 & 37.0 & \textcolor{black!30}{--} & \textcolor{black!30}{--} & 61.2 & \textcolor{black!30}{--} & \textcolor{black!30}{--}\\
\quad Clef-Flash & 9.4 & \textcolor{black!30}{--} & \textcolor{black!30}{--} & \textcolor{black!30}{--} & \textcolor{black!30}{--} & 43.0 & 7.0 & 23.9 & \textcolor{black!30}{--} & \textcolor{black!30}{--} & 45.6 & \textcolor{black!30}{--} & \textcolor{black!30}{--}\\
\quad Winnow-12B & 12 & 16.0\rlap{$^\dagger$} & 14.3\rlap{$^\dagger$} & \textcolor{black!30}{--} & \textcolor{black!30}{--} & \textcolor{black!30}{--} & \textcolor{black!30}{--} & \textcolor{black!30}{--} & \textcolor{black!30}{--} & 81.3 & \textcolor{black!30}{--} & \textcolor{black!30}{--} & \textcolor{black!30}{--}\\
\quad PPLX Decider & 27 & 8.1\rlap{$^\dagger$} & \textbf{19.3} & \textbf{81.1} & \textbf{20.4} & 71.2 & 18.9 & \textbf{41.1} & 44.9 & 82.9 & \textbf{68.6} & 48.6 & \textbf{28.9}\\
\quad Drex-v2 & 35$^a$ & 17.8\rlap{$^\dagger$} & \textcolor{black!30}{--} & \textcolor{black!30}{--} & \textcolor{black!30}{--} & \textcolor{black!30}{--} & \textcolor{black!30}{--} & \textcolor{black!30}{--} & \textcolor{black!30}{--} & 84.1 & \textcolor{black!30}{--} & \textcolor{black!30}{--} & \textcolor{black!30}{--}\\
\bottomrule\end{tabular}
\tabnote{104 model--benchmark runs; the two baseline rows are not counted, and the later controlled transfer reruns are in Table~\ref{tab:replication}, not here. Bold: best model tested per column; --: not run. Failed requests score zero; $^\dagger$\,some requests failed (valid share: BRIGHT default PPLX 44\%, Kev-9B 47\%, Drex-DLM 48\%, Winnow-12B 80\%, Drex-v2 88\%; truncated Winnow-12B 76\%; Amos on SciFact 70\%; Intern-Decision-4B on EnterpriseRAG-Bench 0\%). Columns other than BRIGHT come from earlier project stages with their own input views (on MultiHop-RAG and FinanceBench, candidate text was capped at 128 tokens for PPLX, 256 for Liquid and 512 for Jina and BGE). Typical 95\% half-widths for binary columns: 2--5 points for 443 to 2,167 queries, about 9 points on FinanceBench. Clef-Flash: joint mode; Laya: pointwise. CRAG: PPLX truncated, Jina full text. MuSiQue and Enterpr. were rerun under a frozen protocol (Table~\ref{tab:replication}). $^a$Mixture of experts, 3B active.}
\end{table*}

\begin{table}[t]
\centering
\caption{Which benchmarks does the study use? Sizes, metrics and runs per benchmark}\label{tab:bench}
\scriptsize\setlength{\tabcolsep}{3pt}
\begin{tabular}{@{}l l r l r@{}}\toprule
Benchmark & Content & Queries & Metric & Runs\\
\midrule
BRIGHT~\cite{su2025bright} & Reasoning, 12 domains & 1,384 & nDCG@5 & 13\\
SciFact~\cite{wadden2020scifact} & Scientific claims & 277 & nDCG@5 & 8\\
QASPER~\cite{dasigi2021qasper} & Research papers & 1,502 & FS@5 & 2\\
HotpotQA~\cite{yang2018hotpotqa} & Multi-hop, Wikipedia & 800 & FS@5 & 14\\
MuSiQue~\cite{trivedi2022musique} & Multi-hop, composed & 2,167 & FS@5 & 14\\
2WikiMultihopQA~\cite{ho2020twowiki} & Multi-hop, Wikipedia & 800 & FS@5 & 14\\
MultiHop-RAG~\cite{tang2024multihoprag} & Multi-hop, news & 900 & FS@5 & 5\\
GaRAGe~\cite{sorodoc2025garage} & Grounded QA & 1,397 & Rel@5 & 13\\
EnterpriseRAG-Bench~\cite{sun2026enterpriserag} & Company documents & 443$^a$ & FS@5 & 14\\
FinanceBench~\cite{islam2023financebench} & Financial filings & 109 & Hit@5 & 5\\
CRAG~\cite{yang2024crag} & Web pages & 300 & F1 & 2\\
\midrule Total & & 10,079 & & 104 runs\\
\bottomrule\end{tabular}
\tabnote{Fixed Top-20 list per query (GaRAGe: its supplied pool). FS@5: complete annotated support set in the Top-5; Hit@5: annotated evidence page in the Top-5; Rel@5: share of relevant candidates among the five selected. Runs: Table~\ref{tab:overview}. $^a$443 of 500 requests have resolvable labels; all 500 are run.}
\end{table}

Table~\ref{tab:overview} reports 104 executed model--benchmark runs. It is sparse, not a complete 21-by-11 grid, and not a controlled experiment: runs used different protocols, input representations and candidate-text length caps, and metrics differ by dataset, so scores within a benchmark share a metric but not a fully matched input policy. We use the table to summarize coverage and to pick candidates for deeper study, not to rank models. Blank cells were not run.

\subsection{Why PPLX and Jina were selected}\label{sec:setup}
The broad screen cannot show whether replacement works, because its protocols differ. We therefore fixed one comparison before running the BRIGHT experiments, based on earlier exploratory runs. On a multi-hop task PPLX Decider v1.1 (revision 14c9d90) and Jina Reranker v3.5 were nearly tied (87.8\% versus 88.8\% conditional complete-support retention), on a synthetic enterprise task PPLX was ahead (92.1\% versus 79.7\%), and in an isolated serving test Jina was far faster (p95 86 versus 1,785\,ms): a useful quality--latency trade-off to investigate, and both run on one A100. No completion-rate threshold governed the choice, and default PPLX completes only 44.1\% of BRIGHT requests. We chose Jina over Qwen3-Reranker-4B as the incumbent because it led it in those runs, it reranks a Top-20 set in one request, and it later improved on the first-stage order on all 11 benchmarks.

Carrying PPLX forward was outcome-informed: after default PPLX failed on long inputs and truncation rescued it, it was the only decision model taken through the order, two-stage, budget-sweep and transfer experiments, as a promising truncated decision-model configuration. Default Winnow-12B with a BM25 fallback scores higher on BRIGHT (0.2011 against 0.1928), and our records do not document why it was not carried forward. The case study tests a specific replacement hypothesis, not the model family or every reranker.

The main comparison pits a 27B structured-decision model against a 0.6B dedicated reranker, not a parameter-matched test: the size difference is part of the deployment question. We match the candidate lists and, in controls, the candidate text.

\subsection{Controlled comparison protocol}\label{sec:success}
The controlled study keeps retrieval fixed so we can attribute changes to the reranking step. Each policy receives the same query and Top-20 candidate IDs, so comparisons are paired by request, and the main comparisons measure which Top-5 it returns. We evaluate reliability, ranking quality and serving latency because a replacement must satisfy all three. Invalid outputs and fallback behavior are part of the policy being tested, not reasons to drop a request from the analysis. We do not evaluate the downstream LLM's generated answers.

The controlled experiments use BRIGHT~\cite{su2025bright}: 1,384 reasoning-intensive queries in 12 domains, with the official BM25 Top-20 candidate sets frozen in their original order before labels were joined. EnterpriseRAG-Bench and MuSiQue were rerun later as controlled transfer checks under a frozen protocol, using preserved hybrid (BM25 plus dense) first-stage candidate sets; they are not exact replications of BRIGHT and are not part of the 104 historical runs. A frozen confirmation then used CQADupStack~\cite{hoogeveen2015cqadupstack} from BEIR~\cite{thakur2021beir} (test split, all 12 StackExchange forums, BM25 Top-20 lists, 13,145 queries, of which 20 were reserved for a dry run and 13,125 scored). Before any model output we fixed the settings (6,144 tokens primary, 4,096 secondary), the checks and margin, forums as the resampling unit and a 2,000-query latency subset, and recorded the protocol's SHA-256 hash (f15cb0f6\ldots ce86e).

A new reranker qualifies only if its complete \emph{reranking policy} meets all three requirements. The policy is the checkpoint together with its input handling (truncation and formatting), fallback behavior and execution path, so we evaluate the full reranking setup, including input handling and failures. We separate \emph{reranking capability} (can the model select relevant Top-5 candidates once its input is handled sensibly?) from \emph{deployment suitability} (does the whole policy pass all three requirements?). Passing the first does not imply passing the second.
\begin{itemize}
\item \textbf{Reliability:} at least 99\% of requests return a valid Top-5. The strict reading counts only selections the model itself returns. The lenient reading also counts a fallback to the BM25 Top-5, because users still receive a Top-5. A model that fails most requests can meet the lenient reading, so the budget sweep applies the strict reading and reports the lenient one as a sensitivity check.
\item \textbf{Ranking quality:} the policy beats the BM25 Top-5 (the lower end of its 95\% interval is above zero) and stays within the maximum acceptable loss against full-text Jina (non-inferiority: the lower end of its 95\% interval for the difference is above $-\delta$, with $\delta=0.02$ nDCG@5). The margin is about 40\% of Jina's gain over the BM25 Top-5 (0.046 on the 607 judged queries, see below) and was set after the first results; Section~\ref{sec:rq3} shows which conclusions survive a stricter $\delta=0.01$.
\item \textbf{Serving:} p95 latency (the 95th-percentile request latency) at most twice Jina's on the same GPU. An observed ratio above 2 is not approved; if its bootstrap interval spans 2, the evidence is inconclusive, not an established violation.
\end{itemize}
These requirements are illustrative engineering choices, not industry-standard service-level targets; a deployment would set its own latency and GPU-memory budgets. GPU memory is not a gate here; we report it separately.

\textbf{Score every request.} A valid selection is five distinct candidates from the original Top-20; an error or a repeated, out-of-list or missing ID is a failure. Without a fallback, a failed request receives zero quality credit instead of disappearing from the average; with a fallback, it is scored by the BM25 Top-5 it serves. Either way every evaluated request counts. Formally, let $V_\pi(q)=1$ if policy $\pi$ returns a valid selection $S_\pi(q)$ for query $q$, and let $m(q,S)$ be its quality. The \emph{all-request score} is
\begin{equation}\label{eq:oper}
\Qop(\pi)=\frac{1}{N}\sum_{q=1}^N V_{\pi}(q)\,m\big(q,S_{\pi}(q)\big).
\end{equation}
The \emph{valid-only score} averages $m$ over completed requests only. Dropping failed requests favors a model that is evaluated only on the cases it can handle, and it hides the relevance a competitor finds on the requests the model fails. We call this bias from excluding failed requests the \emph{valid-only trap}; it can bias the comparison, and it does not show that the model cannot be improved (Fig.~\ref{fig:decomp}). A \emph{fallback} policy serves the BM25 Top-5 whenever the model fails, as a production system would.

\textbf{Ranking quality.} On BRIGHT we use nDCG@5 normalized by the best score achievable from all of a query's annotated positives. The judgments are incomplete: only 607 of the 1,384 candidate sets contain a judged positive, and on the rest every policy scores zero, so they cannot separate policies and a failure there costs nothing. ``No judged positive'' does not mean ``no relevant candidate'', since unjudged candidates may still be relevant. Those 777 zeros shrink every all-request score and gap to $607/1{,}384=0.44$ of its value on the 607 judged queries: multiply BRIGHT nDCG@5 values and differences by 2.28 (full-text Jina 0.1844 becomes 0.420, the BM25 Top-5 0.133 becomes 0.303, and the 0.02 margin becomes 0.046).

\textbf{Statistics.} Queries within a BRIGHT domain are related, so 95\% intervals resample whole domains (40,000 bootstrap draws~\cite{efron1993bootstrap} over 12 domains), and an exact sign-flip test over all $2^{12}$ domain sign patterns~\cite{smucker2007significance} gives $p$-values, adjusted with Holm's method within each family of contrasts. For the budget sweep we also report selection-adjusted lower bounds over all six PPLX budgets: the same domain resampling, with each budget's deviation divided by its bootstrap standard deviation and the 2.5th percentile of the smallest value across budgets as a common critical value (simultaneous bound), or the 0.025/6 tail (Bonferroni, more conservative). Reliability uses the observed rate; quality and serving use the 95\% bounds above. The adjusted, exact-validity and two-setting bounds are added sensitivity checks. Tables report point estimates; intervals appear in the text. With 12 domains the interval and test can disagree and intervals can be optimistic~\cite{cameron2008cluster}, so an effect whose interval excludes zero but whose adjusted $p$ exceeds 0.05 is reported as not established.

\textbf{Input handling.} Dedicated rerankers can usually read a whole candidate set, but several decision models cannot. Jina's default call truncates only 48 of 27,680 BRIGHT candidate texts (0.17\%). The decision models accept 8,192 tokens, but the median BRIGHT candidate set is 8,097 tokens, so 49\% of sets overflow before any prompt is added. We test a straightforward repair: reserve room for the question and the prompt, then allocate the same token budget to each of the $K=20$ candidates (\emph{equal-prefix truncation}), so that one long candidate cannot crowd out the others, at the cost of possibly discarding evidence a candidate needs. The model receives the prefix of each candidate's text; the candidate IDs never change.
\begin{equation}
 b(q)=\max\!\left(0,\left\lfloor\frac{L-h(q)-\rho}{K}\right\rfloor\right),\quad \rho=128, \label{eq:budget}
\end{equation}
Here $L$ is the input-token budget: 8,192 is the decision models' limit, and Section~\ref{sec:sweep} sweeps it from 1,024 to 8,192. Each candidate keeps its first $b(q)$ tokens, counted in the model's own tokenizer (the PPLX tokenizer for Jina in matched-text runs). The prompt overhead $h(q)$ is measured on the exact request with empty candidate texts. The question is never truncated, unused budget is not redistributed, and if the reassembled request still overflows, we reduce the per-candidate budget one token at a time until it fits. When even that cannot produce a request, the request fails, for Jina and PPLX alike; this \emph{builder failure} matters only at small $L$. The rule uses no labels or outcomes.

Different truncation policies can give models different information. To separate ranking decisions from the effect of reading more text, we compare against a dedicated reranker that sees the same candidate texts: in the \emph{matched-text control}, Jina scores exactly the strings given to the decision model. This holds the visible text constant, not model size or compute.

\textbf{Serving.} We time one request at a time (no batching and no concurrent load; we call these serial GPU measurements) on an NVIDIA A100-SXM4-80GB. Latency covers request construction, inference, parsing, ID mapping and packaging the Top-5, but not model loading. Percentiles cover completed requests; all-request latency counts a failed request as its time to failure plus the BM25 fallback. The budget sweep was timed on all 1,384 queries in two sessions of two passes each, every session with its own full-text Jina anchor (p95 about 1,210\,ms), and intervals on each p95 ratio come from a bootstrap over query IDs (10,000 draws). The earlier comparisons, and the few sweep rows marked in Table~\ref{tab:sweep}, use a cohort of 120 BRIGHT queries fixed by a seed before any serving result was seen and timed three times in dedicated-GPU sessions; the other decision models were timed in earlier sessions on shared nodes. All runs used a shared host. This is an isolated latency comparison, not a throughput or concurrency study.

\textbf{Planned and post-hoc analyses.} All-request scoring, the BM25 and Jina contrasts and the serving cohort were fixed in advance. The margin $\rho$ was chosen after PPLX's default failures; the fallback, $\delta$, the Holm families and the label-coverage analysis came afterward. The follow-up experiments and the transfer check, including its budget grid and timing, ran under protocols frozen before their outcomes; the transfer check used preserved hybrid first-stage candidate sets because BM25-only sets were unavailable. The budget sweep was requested after the 8,192-token results: its grid and checks were fixed before it ran, but which budget qualifies was not, so its headline is a finding to confirm.

\section{Evaluation}
The broad screen suggests that decision models can be competitive on some retrieval tasks, but it cannot establish a replacement because the runs used different protocols. We therefore compare policies on the same candidate lists, organized around the main obstacles to replacement: invalid outputs (Section~\ref{sec:rq1}), input handling and multi-stage selection (Section~\ref{sec:rq2}), input length (Section~\ref{sec:sweep}), serving latency (Section~\ref{sec:rq3}) and transfer to other workloads (Section~\ref{sec:overview}).

\textbf{Broad-screen findings.} In Table~\ref{tab:overview}, Jina's point estimate exceeds the first-stage order on all 11 benchmarks (on CRAG its interval includes zero). PPLX is competitive when inputs fit: it led five of the other ten benchmark columns and the truncated BRIGHT column under their historical input policies, though several runs capped candidate text at 128 tokens. Failures recur (daggers in Table~\ref{tab:overview}); Intern-Decision-4B hits its context limit on all 500 EnterpriseRAG-Bench requests.

\subsection{Does default reranking work reliably?}\label{sec:rq1}
\textbf{The default PPLX configuration fails on many long candidate sets: it returns a valid Top-5 for only 44.1\% of BRIGHT queries} (Table~\ref{tab:original}), so scores computed only on successful requests give a misleading picture. A model that fails on more than half its requests cannot serve as a drop-in reranker, even if its completed requests rank well: on completed requests PPLX beats Jina by $+0.0371$, but over all requests it falls below Jina by $0.1031$.

\begin{table}[t]
\centering
\caption{Does default one-call reranking survive counting failures? Completed-only versus all-request scores (BRIGHT, $N=1{,}384$)}\label{tab:original}
\scriptsize\setlength{\tabcolsep}{3pt}
\begin{tabular}{@{}lrrrr@{}}\toprule
 & Done & & \multicolumn{2}{c}{$\Delta$ nDCG@5 vs Jina}\\\cmidrule(l){4-5}
Policy & (\%) & nDCG@5 & completed only & all requests\\
\midrule
BM25 Top-5 & 100 & 0.1330 & -- & $-.0514$\\
Jina v3.5 & 100 & \textbf{0.1844} & -- & --\\
PPLX Decider, default & 44.1 & 0.0813 & $\mathbf{+.0371}$ & $\mathbf{-.1031}$\\
\quad + BM25 fallback & 100 & 0.1687 & -- & $-.0157$\\
Winnow-12B, default & 79.6 & 0.1599 & $+.0325$ & $-.0245$\\
\quad + BM25 fallback & 100 & 0.2011 & -- & $+.0168$\\
Drex-v2, default & 87.6 & 0.1781 & $+.0249$ & $-.0062$\\
\quad + BM25 fallback & 100 & 0.2005 & -- & $+.0161$\\
\bottomrule\end{tabular}
\tabnote{Completed only: difference from Jina on the requests the policy completed. All requests: a failed request scores zero; with fallback, it receives the BM25 Top-5 instead. Other models' BRIGHT scores are in Table~\ref{tab:overview}.}
\end{table}

\begin{figure}[t]
\centering\includegraphics[width=\columnwidth]{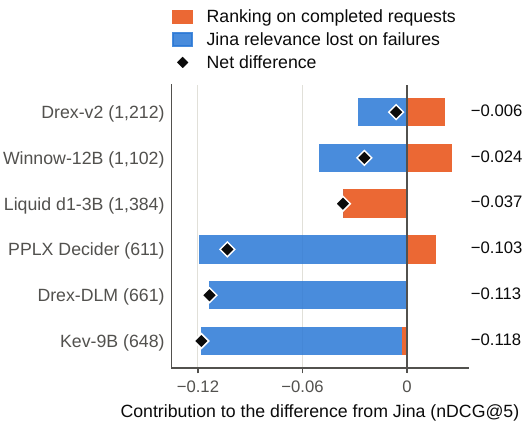}
\caption{Counting failed requests removes the apparent quality gain. Each bar splits a policy's all-request difference from Jina into its ranking gain on the requests it completed (orange) and the relevance Jina finds on the requests it failed (blue). Diamonds and right-hand values give the net difference; parentheses give completed requests out of 1,384.}
\label{fig:decomp}
\end{figure}

\textbf{Jina scores higher on the queries for which PPLX could not return a valid ranking.} Jina gains $+0.0514$ over the BM25 Top-5 (interval $[+0.0383,+0.0663]$, all 12 domains, Holm $p=0.002$); that gain is what a replacement has to preserve. The failed requests are not random: Jina scores 0.214 on the 773 candidate sets PPLX fails but only 0.147 on the 611 it completes, and label coverage does not explain the gap (45\% vs 43\% contain a judged positive). The failures are long inputs. In Fig.~\ref{fig:decomp}, PPLX's completed requests add $+0.0164$ to its difference from Jina, while the relevance Jina finds on PPLX's failed requests subtracts $0.1195$.

\textbf{A fallback to the BM25 Top-5 helps some models but not PPLX.} Serving the BM25 Top-5 on failed requests lifts default Winnow-12B to 0.2011 and Drex-v2 to 0.2005, ahead of Jina by $+0.0168$ and $+0.0161$ and within the quality budget. PPLX with fallback still trails Jina by 0.016 with a lower bound of $-0.037$, outside the budget, so default PPLX does not meet the ranking-quality check. The failures come from input length, so the next question is whether input handling can repair them.

\subsection{Can input handling make decision-model reranking reliable and accurate?}\label{sec:rq2}
We test context length (truncation prevents overflow but may discard text), candidate text (a Jina control that reads the same text), candidate order (reversed and shuffled) and two-stage reranking, which we also apply to Jina to separate the model from the procedure.

\textbf{Truncating candidate text lets PPLX process nearly all BRIGHT queries. It does not improve every decision model: Kev-9B and Drex-DLM become more reliable but remain weak rerankers.} With equal-prefix truncation at the models' 8,192-token limit, completed requests rise from 611 to 1,380 for PPLX and to all 1,384 for Kev-9B and Drex-DLM (Table~\ref{tab:budget}). Truncated PPLX scores 0.1928, above the BM25 Top-5 by $+0.0598$ (interval $[+0.0377,+0.0840]$, Holm $p=0.004$) and within the quality budget of full-text Jina ($+0.0084$, lower bound $-0.0124$), though not shown equivalent. Against Jina reading the same truncated text, PPLX leads by $+0.0163$ (Holm $p=0.10$), which we do not claim as a difference. Truncated Kev-9B and Drex-DLM score 0.121 and 0.119, below the BM25 Top-5.

\begin{table}[t]
\centering
\caption{Does truncating candidate text make decision models competitive? Equal-prefix truncation to 8,192 tokens on BRIGHT}\label{tab:budget}
\scriptsize\setlength{\tabcolsep}{3.2pt}
\begin{tabular}{@{}l rr rr r rr@{}}\toprule
 & \multicolumn{2}{c}{Done (\%)} & \multicolumn{2}{c}{nDCG@5} & Jina & \multicolumn{2}{c}{$\Delta$ of truncated vs}\\\cmidrule(lr){2-3}\cmidrule(lr){4-5}\cmidrule(l){7-8}
Decision model & default & trunc. & default & trunc. & same text & Jina & BM25 Top-5\\
\midrule
PPLX Decider & 44.1 & 99.7 & 0.0813 & \textbf{0.1928} & 0.1765 & $\mathbf{+.0163}$ & $\mathbf{+.0598}$\\
Kev-9B & 46.8 & 100 & 0.0663 & 0.1213 & 0.1779 & $-.0566$ & $-.0116$\\
Drex-DLM & 47.8 & 100 & 0.0710 & 0.1187 & 0.1770 & $-.0582$ & $-.0142$\\
Winnow-12B & 79.6 & 75.5 & 0.1599 & 0.1428 & 0.1844$^a$ & $-.0415$ & $+.0099$\\
\bottomrule\end{tabular}
\tabnote{Jina same text: Jina scores exactly the truncated text the decision model read. $^a$\,No same-text control exists for Winnow-12B, whose truncated run used a different server build; its column shows full-text Jina.}
\end{table}

\textbf{PPLX's ranking score changes when the same candidates arrive in a different order.} A listwise model may benefit from receiving better candidates first~\cite{tang2024found,zheng2024mcq}. We rearrange the same Top-20 candidates and rerun each model, in reversed order and five seeded random orders on all 1,384 queries; the candidate IDs and texts stay fixed. A new order changed PPLX's Top-5 for 97\% of queries and Jina's for 78\%. Jina's average score barely moves ($-0.0004$), but PPLX scores 0.013 higher in BM25 order than in shuffled orders, in all 12 domains (Holm $p=0.002$): some of its measured quality comes from the candidate order. In shuffled orders truncated PPLX still beats the BM25 Top-5, but trails Jina by 0.005 with a lower bound of $-0.024$, outside the quality budget (Table~\ref{tab:followup}).

\textbf{Multi-stage reranking improves PPLX's ranking but costs three inference calls.} The policy first ranks two groups of ten candidates (truncated only if needed) and keeps five from each; a third call selects the final five from those ten. This policy completes 1,381 of 1,384 requests and scores 0.2108, ahead of Jina by $+0.0264$ (Holm $p=0.010$) and ahead of single-call PPLX by $+0.0180$. It is also more robust to order: averaged over six shuffled orders it leads Jina by $+0.0198$ (Holm $p=0.039$) and loses only 0.006 relative to BM25 order.

\textbf{Does multi-stage reranking help any model, or PPLX in particular?} Two-stage ranking is common for LLM rankers~\cite{sun2023rankgpt,zhuang2024setwise}, so the gain might come from the procedure or from the amount of text each model reads. We therefore ran Jina through the same windows and survivor rule in all seven orders. We detect no significant improvement for Jina ($+0.0025$ over single-call Jina), while two-stage PPLX beats two-stage full-text Jina by $+0.0239$ (Holm $p=0.016$) in BM25 order. In a second control, Jina received exactly PPLX's per-window strings and selected its own five survivors per window, so its final ten candidates rarely coincide with PPLX's (8 of 1,381 requests). Two-stage PPLX still leads, by $+0.0252$ in BM25 order (Holm $p=0.008$) and \textbf{$+0.0213$ over six shuffled orders (Holm $p=0.020$)}, positive in 10 of 12 domains (Table~\ref{tab:followup}). The sweep's two-stage Jina reuses an earlier trace that scores 0.1790, while this matched-text control scores 0.1855; an audit found that the two runs fed Jina differently serialized window text (Top-5 lists differ on 1,069 queries, scores on 197) and that tokenizer, dtype and batch settings were not recorded, so we report both, and the control gives the smaller PPLX lead ($+0.025$, not $+0.032$). This result is specific to the evaluated PPLX and Jina configurations.

\begin{table}[t]
\centering
\caption{Do multi-stage reranking and its controls hold up? Follow-up experiments on BRIGHT}\label{tab:followup}
\scriptsize\setlength{\tabcolsep}{3pt}
\begin{tabular}{@{}l r rr@{}}\toprule
Contrast (nDCG@5) & $\Delta$ & $p$ & Holm $p$\\
\midrule
\addlinespace[2pt]\multicolumn{4}{@{}l}{\textit{Candidate order (row 3: mean over six shuffled orders)}}\\[1pt]
\quad PPLX: BM25 $-$ shuffled order & $\mathbf{+.0131}$ & $<$.001 & 0.002\\
\quad Jina: BM25 $-$ shuffled order & $-.0004$ & 0.848 & --\\
\quad PPLX truncated $-$ Jina & $-.0051$ & 0.620 & --\\
\addlinespace[2pt]\multicolumn{4}{@{}l}{\textit{Two-stage PPLX (two windows of 10, then 10 to 5)}}\\[1pt]
\quad Two-stage $-$ Jina & $\mathbf{+.0264}$ & 0.003 & 0.010\\
\quad Two-stage $-$ PPLX truncated & $\mathbf{+.0180}$ & 0.020 & 0.039\\
\quad Two-stage $-$ Jina, shuffled & $\mathbf{+.0198}$ & 0.020 & 0.039\\
\quad Two-stage: BM25 $-$ shuffled order & $+.0062$ & 0.051 & --\\
\addlinespace[2pt]\multicolumn{4}{@{}l}{\textit{Same two-stage procedure for Jina$^b$}}\\[1pt]
\quad PPLX two-stage $-$ Jina two-stage & $\mathbf{+.0239}$ & 0.005 & 0.016\\
\quad \quad same, shuffled & $+.0193$ & 0.029 & 0.059\\
\quad PPLX two-stage $-$ Jina, same text$^c$ & $\mathbf{+.0252}$ & 0.004 & 0.008\\
\quad \quad same, shuffled & $\mathbf{+.0213}$ & 0.020 & 0.020\\
\quad Two-stage Jina $-$ Jina & $+.0025$ & 0.232 & 0.232\\
\bottomrule\end{tabular}
\tabnote{Bold: Holm-adjusted $p<0.05$. Protocols were frozen before outcomes; whole-domain bootstrap and exact sign-flip test; Holm adjustment within each pre-declared family: the $^b$\,control group (four contrasts; the omitted one is two-stage PPLX minus two-stage Jina on the single-call truncated strings, $+0.0318$, Holm $p=0.004$) and the identical-text pair ($^c$\,Jina read exactly the per-window strings PPLX received). Shuffled: mean over six candidate orders. --: outside a Holm family.}
\end{table}

Truncation to 8,192 tokens was one arbitrary setting, and serving cost grows with input length. How much input does the model actually need?

\subsection{How much input does a decision model need?}\label{sec:sweep}
\begin{figure}[t]
\centering\includegraphics[width=\columnwidth]{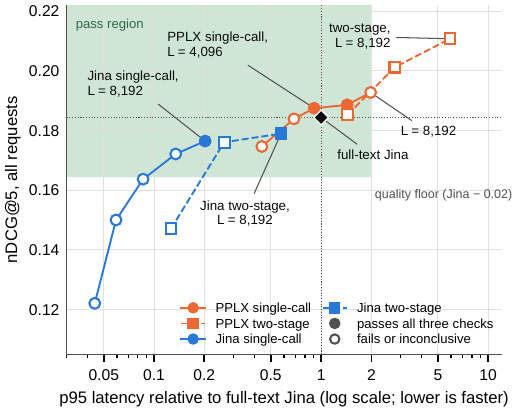}
\caption{Quality against latency across input budgets (BRIGHT; each point is one budget, from 2,048 to 8,192 tokens). The shaded region meets the quality and latency checks; filled markers also meet the reliability check. Single-call PPLX at 4,096 and 6,144 tokens meets the checks, but only 4,096 is near parity with full-text Jina (6,144 is slower); a tuned Jina is faster; two-stage PPLX is better but slower.}
\label{fig:frontier}
\end{figure}

We swept the input budget $L$ in Eq.~(\ref{eq:budget}) from 1,024 to 8,192 tokens for single-call PPLX and Jina on identical truncated text, and for the two-stage policies at three budgets, scoring all 1,384 queries and timing nearly every row on the same queries (Table~\ref{tab:sweep}, Fig.~\ref{fig:frontier}). Strict scoring applies throughout: a failed request counts as zero and the model must itself return a valid Top-5 for 99\% of requests.

\begin{table*}[t]
\centering
\caption{How much input does a decision model need? Input-token budget sweep on BRIGHT (1,384 queries; strict scoring: a failed request scores zero)}\label{tab:sweep}
\footnotesize\setlength{\tabcolsep}{2.6pt}
\begin{tabular}{@{}r r rrc rrc r rrc rrc@{}}\toprule
 & & \multicolumn{3}{c}{Single call: PPLX} & \multicolumn{3}{c}{Single call: Jina} & & \multicolumn{3}{c}{Two-stage: PPLX} & \multicolumn{3}{c}{Two-stage: Jina}\\\cmidrule(lr){3-5}\cmidrule(lr){6-8}\cmidrule(lr){10-12}\cmidrule(l){13-15}
Budget $L$ & Valid & nDCG@5 & p95$\times$ & R\,Q\,S & nDCG@5 & p95$\times$ & R\,Q\,S & Valid & nDCG@5 & p95$\times$ & R\,Q\,S & nDCG@5 & p95$\times$ & R\,Q\,S\\
\midrule
1,024 & 35.0 & 0.0458 & 0.23$^\dagger$ & \textcolor{cFail}{$\times$} \textcolor{cFail}{$\times$} \textcolor{cPass!70!black}{$\checkmark$} & 0.0242 & 0.03$^\dagger$ & \textcolor{cFail}{$\times$} \textcolor{cFail}{$\times$} \textcolor{cPass!70!black}{$\checkmark$} & \textcolor{black!30}{--} & \textcolor{black!30}{--} & \textcolor{black!30}{--} & \textcolor{black!30}{--} & \textcolor{black!30}{--} & \textcolor{black!30}{--} & \textcolor{black!30}{--}\\
2,048 & 94.9 & 0.1747 & 0.44 & \textcolor{cFail}{$\times$} \textcolor{cFail}{$\times$} \textcolor{cPass!70!black}{$\checkmark$} & 0.1221 & 0.04$^\dagger$ & \textcolor{cFail}{$\times$} \textcolor{cFail}{$\times$} \textcolor{cPass!70!black}{$\checkmark$} & 97.1 & 0.1854 & 1.4$^\dagger$ & \textcolor{cFail}{$\times$} \textcolor{cPass!70!black}{$\checkmark$} \textcolor{black!55}{?} & 0.1471 & 0.13$^\dagger$ & \textcolor{cFail}{$\times$} \textcolor{cFail}{$\times$} \textcolor{cPass!70!black}{$\checkmark$}\\
3,072 & 98.7 & 0.1839 & 0.69 & \textcolor{cFail}{$\times$} \textcolor{cPass!70!black}{$\checkmark$} \textcolor{cPass!70!black}{$\checkmark$} & 0.1500 & 0.06$^\dagger$ & \textcolor{cFail}{$\times$} \textcolor{cFail}{$\times$} \textcolor{cPass!70!black}{$\checkmark$} & \textcolor{black!30}{--} & \textcolor{black!30}{--} & \textcolor{black!30}{--} & \textcolor{black!30}{--} & \textcolor{black!30}{--} & \textcolor{black!30}{--} & \textcolor{black!30}{--}\\
4,096 & 99.2 & \textbf{0.1875} & \textbf{0.91} & \textcolor{cPass!70!black}{$\checkmark$} \textcolor{cPass!70!black}{$\checkmark$} \textcolor{cPass!70!black}{$\checkmark$} & 0.1637 & 0.09$^\dagger$ & \textcolor{cPass!70!black}{$\checkmark$} \textcolor{cFail}{$\times$} \textcolor{cPass!70!black}{$\checkmark$} & 99.2 & 0.2013 & 2.8 & \textcolor{cPass!70!black}{$\checkmark$} \textcolor{cPass!70!black}{$\checkmark$} \textcolor{cFail}{$\times$} & 0.1761 & 0.26$^\dagger$ & \textcolor{cPass!70!black}{$\checkmark$} \textcolor{cFail}{$\times$} \textcolor{cPass!70!black}{$\checkmark$}\\
6,144 & 99.6 & \textbf{0.1886} & \textbf{1.4} & \textcolor{cPass!70!black}{$\checkmark$} \textcolor{cPass!70!black}{$\checkmark$} \textcolor{cPass!70!black}{$\checkmark$} & 0.1721 & 0.13 & \textcolor{cPass!70!black}{$\checkmark$} \textcolor{cFail}{$\times$} \textcolor{cPass!70!black}{$\checkmark$} & \textcolor{black!30}{--} & \textcolor{black!30}{--} & \textcolor{black!30}{--} & \textcolor{black!30}{--} & \textcolor{black!30}{--} & \textcolor{black!30}{--} & \textcolor{black!30}{--}\\
8,192 & 99.7 & 0.1928 & 2.0 & \textcolor{cPass!70!black}{$\checkmark$} \textcolor{cPass!70!black}{$\checkmark$} \textcolor{black!55}{?} & \textbf{0.1765} & \textbf{0.20} & \textcolor{cPass!70!black}{$\checkmark$} \textcolor{cPass!70!black}{$\checkmark$} \textcolor{cPass!70!black}{$\checkmark$} & 99.8 & 0.2108 & 5.9$^\dagger$ & \textcolor{cPass!70!black}{$\checkmark$} \textcolor{cPass!70!black}{$\checkmark$} \textcolor{cFail}{$\times$} & \textbf{0.1790} & \textbf{0.58} & \textcolor{cPass!70!black}{$\checkmark$} \textcolor{cPass!70!black}{$\checkmark$} \textcolor{cPass!70!black}{$\checkmark$}\\
\bottomrule\end{tabular}
\tabnote{Full-text Jina: nDCG@5 0.1844, p95 1,211\,ms (ratio 1.0). p95 columns give the ratio to that value (all 1,384 queries, two sessions with two passes each, one request at a time, A100, same-session Jina anchor; $^\dagger$\,rows use a 120-query cohort timed three times, so they are not directly comparable with the other rows). Valid (\%): requests for which the truncated request builder succeeds and the model returns five valid IDs; Jina rows have the same valid rate as the PPLX row beside them, except two-stage Jina at 8,192 (99.7). R: observed valid rate $\ge$99\%. Q: beats the BM25 Top-5 and lower 95\% bound vs full-text Jina above $-0.02$. S: p95 ratio interval at or below 2 ($\checkmark$, pass), spanning 2 (?, inconclusive) or above 2 ($\times$, violation). A check mark means the rule is met, not that reliability is statistically established or that the setting is proven for deployment; the budgets were chosen exploratorily. Bold: passes all three. Jina single-call at 8,192 takes its quality from the earlier frozen run (the sweep re-run hit 32 client errors); its timing comes from the full-population sessions.}
\end{table*}

\textbf{PPLX needs far less input than Jina to stay near full-text quality.} PPLX reaches 0.1839 with 3,072 tokens per request, within 0.0005 of full-text Jina, and 0.1875 at 4,096. Jina on the same text falls to 0.1500 and 0.1637 and needs 8,192 tokens to come within 0.008. At equal budget PPLX leads Jina by $+0.053$ at 2,048 tokens, shrinking to $+0.016$ at 8,192. With Holm adjustment the lead is supported at 2,048 and 3,072 tokens ($p=0.002$, $0.012$) but not at 4,096 and above ($p=0.067$ to $0.095$).

\textbf{Latency grows with the budget, so only a narrow band qualifies.} PPLX's p95 rises from 0.44 times full-text Jina's at 2,048 tokens to 0.91 at 4,096, 1.43 at 6,144 and 1.98 at 8,192. \textbf{Single-call PPLX passes all three checks at 4,096 and at 6,144 tokens}, and no other PPLX policy does. At 4,096 tokens it returns 99.2\% valid Top-5s, scores nDCG@5 0.1875 ($+0.0031$ against full-text Jina, lower bound $-0.0146$) and has a p95 of 1,101\,ms, 0.91 times Jina's (interval $[0.86,1.02]$). At 6,144 tokens the figures are 99.6\%, 0.1886 ($+0.0042$, lower bound $-0.0145$) and 1,735\,ms, 1.43 times ($[1.35,1.61]$). The 8,192-token budget is inconclusive (1.98, $[1.87,2.22]$). Both meet the twice-Jina limit, but neither shows lower latency: the 4,096 interval includes 1, and 6,144 is clearly slower. Below 4,096 tokens the builder cannot fit 1.3\% (3,072) to 5.1\% (2,048) of requests, failing reliability.

\textbf{Parity is an average over domains.} PPLX's latency barely varies across the 12 BRIGHT domains (p95 1,049 to 1,104\,ms at 4,096 tokens) because the budget fixes its input length, whereas Jina's follows candidate text length (158\,ms to 3.5\,s). The per-domain p95 ratio therefore runs from 0.31 (Earth Science) to 6.8 (Pony) and exceeds 2 in three domains (Biology, Pony, Sustainable Living), so a workload with short candidate texts would not meet the latency check.

\textbf{Staging adds quality at a price.} Two-stage PPLX at 4,096 tokens scores 0.2013 ($+0.0169$ against full-text Jina) at 2.8 times Jina's p95 (interval $[2.6,3.1]$), clearly above the limit; at 8,192 tokens it reaches 0.2108 at 5.9 times. Two-stage Jina at 8,192 tokens passes all three checks (0.1790, 0.58 times).

\textbf{Caveat.} This sweep is exploratory. We tested six budgets (1,024 to 8,192 tokens) and found the two adjacent passing settings after seeing the results, so they are candidates for confirmation, which the frozen tests on other workloads (Section~\ref{sec:overview}) support only in part. Observed compliance is not statistical support. Adjusted for the six budgets, the quality lower bounds are $-0.0185$ (4,096) and $-0.0189$ (6,144) under a simultaneous bootstrap, above the $-0.02$ margin, but $-0.0205$ and $-0.0208$ under Bonferroni; the exact 95\% lower bounds on the valid rate are 98.6\% and 99.1\%, so only 6,144 clears the 99\% check with support. Some rows use 120-query timing, and the 4,096-token ratio interval reaches parity. The sweep's Jina at 8,192 tokens reuses the frozen run. A pass means a setting meets the checks, not that it is better.

\subsection{Does the quality improvement justify the serving cost?}\label{sec:rq3}
\textbf{Better ranking quality often requires more inference time, and the gain alone does not make PPLX the better deployment choice.} The best-ranking policy, three-call PPLX at 8,192 tokens, has a p95 of 6,700\,ms, \textbf{5.9 times} Jina's (interval $[3.8,8.5]$); two earlier dedicated sessions measured 6.0 and 5.8 times. Other decision models fare no better: Liquid d1-3B, the only one fully reliable by default, has 3.0 times Jina's p95, and Winnow-12B with fallback 4.6 (indicative only). A 27B model can be slower than a 0.6B reranker even when it outputs only a list of IDs.

\textbf{Tune the Jina baseline before replacing it.} Jina on the truncated 8,192-token text scores 0.1765, below full-text Jina by $0.0079$ (interval $[-0.0171,+0.0008]$) and well above the BM25 Top-5, at a p95 of \textbf{245\,ms (0.20 times full-text Jina, interval $[0.19,0.23]$)} and 1.9\,GiB. It passes all three checks with only a changed input policy, and in the same timing sessions it is 4.5 to 7 times faster than the passing PPLX configurations, at 0.011 to 0.012 lower nDCG@5. Shortening Jina to 4,096 tokens costs 0.0207 (lower bound $-0.0406$) and fails the quality check, and it hurts on EnterpriseRAG-Bench (Section~\ref{sec:overview}), so the best tuned option is workload-specific.

\textbf{Memory and reuse.} PPLX occupies about 50\,GiB at peak against 4.3\,GiB for full-text Jina, so even the passing PPLX policies need more than 11 times Jina's memory.

\textbf{Summing up (Table~\ref{tab:decision}).} On BRIGHT, two single-call PPLX settings (4,096 and 6,144 tokens) meet our three checks, as non-inferior to Jina within the 0.02 margin rather than shown equivalent. Neither shows lower latency, both need more than 11 times the memory, and a tuned Jina is faster still. The best-ranking policies, two-stage PPLX and Winnow-12B with fallback, are the slowest. The verdict also depends on the quality margin: under a stricter $\delta=0.01$ no single-call policy passes (PPLX's lower bounds at 4,096, 6,144 and 8,192 tokens are $-0.0146$, $-0.0145$ and $-0.0124$), and the policies that clear quality fail serving.

\begin{table*}[t]
\centering
\caption{Which policies pass the three replacement checks? Summary of BRIGHT policies against the dedicated reranker (1,384 queries)}\label{tab:decision}
\footnotesize\setlength{\tabcolsep}{3pt}
\begin{tabular}{@{}l r r r r r r ccc l@{}}\toprule
 & Valid & & $\Delta$ vs & p95 & p95 & Mem. & \multicolumn{3}{c}{Checks} & \\\cmidrule(lr){8-10}
Reranking policy & (\%) & nDCG@5 & Jina & (ms) & ratio & (GiB) & Rel. & Qual. & Serv. & Takeaway\\
\midrule
\addlinespace[2pt]\multicolumn{11}{@{}l}{\textit{Baseline and dedicated reranker (Jina v3.5)}}\\[1pt]
BM25 Top-5 (no reranker) & 100 & 0.1330 & $-.0514$ & -- & -- & -- & \textcolor{cPass!70!black}{$\checkmark$} & \textcolor{black!40}{--} & \textcolor{black!40}{--} & Baseline to beat\\
Jina v3.5, full text (dedicated) & 100 & 0.1844 & ref. & 1,211 & 1.0 & 4.3 & \textcolor{cPass!70!black}{$\checkmark$} & \textcolor{cPass!70!black}{$\checkmark$} & \textcolor{cPass!70!black}{$\checkmark$} & Dedicated reranker\\
Jina v3.5, truncated to 8,192 & 99.7 & 0.1765 & $-.0079$ & 245 & 0.20 & 1.9 & \textcolor{cPass!70!black}{$\checkmark$} & \textcolor{cPass!70!black}{$\checkmark$} & \textcolor{cPass!70!black}{$\checkmark$} & Tuned Jina; passes all\\
\addlinespace[2pt]\multicolumn{11}{@{}l}{\textit{Main case, PPLX Decider: default input, then input budget and multi-stage reranking}}\\[1pt]
PPLX, default input & 44.1 & 0.0813 & $-.1031$ & -- & -- & -- & \textcolor{cFail}{$\times$} & \textcolor{cFail}{$\times$} & \textcolor{black!40}{--} & Fails 56\% of requests\\
PPLX, default + BM25 fallback & 100 & 0.1687 & $-.0157$ & -- & -- & -- & \textcolor{cPass!70!black}{($\checkmark$)} & \textcolor{cFail}{$\times$} & \textcolor{black!40}{--} & Non-inferiority not shown\\
\textbf{PPLX, single call, 4,096} & \textbf{99.2} & \textbf{0.1875} & $\mathbf{+.0031}$ & \textbf{1,101} & \textbf{0.91} & 49.4 & \textcolor{cPass!70!black}{$\checkmark$} & \textcolor{cPass!70!black}{$\checkmark$} & \textcolor{cPass!70!black}{$\checkmark$} & Passes all three checks\\
\textbf{PPLX, single call, 6,144} & \textbf{99.6} & \textbf{0.1886} & $\mathbf{+.0042}$ & \textbf{1,735} & \textbf{1.4} & 49.7 & \textcolor{cPass!70!black}{$\checkmark$} & \textcolor{cPass!70!black}{$\checkmark$} & \textcolor{cPass!70!black}{$\checkmark$} & Passes all three; slower\\
PPLX, single call, 8,192 & 99.7 & 0.1928 & $+.0084$ & 2,397 & 2.0 & 50.1 & \textcolor{cPass!70!black}{$\checkmark$} & \textcolor{cPass!70!black}{$\checkmark$} & \textcolor{black!55}{?} & At the latency limit\\
\quad same, shuffled order & 99.7 & 0.1797 & $-.0051$ &  &  &  & \textcolor{cPass!70!black}{$\checkmark$} & \textcolor{cFail}{$\times$} & \textcolor{black!55}{?} & Relies on candidate order\\
PPLX, two-stage (3 calls), 4,096 & 99.2 & 0.2013 & $+.0169$ & 3,352 & 2.8 & 49.4 & \textcolor{cPass!70!black}{$\checkmark$} & \textcolor{cPass!70!black}{$\checkmark$} & \textcolor{cFail}{$\times$} & Better quality; too slow\\
PPLX, two-stage (3 calls), 8,192 & 99.8 & 0.2108 & $+.0264$ & 6,700 & 5.9 & 50.1 & \textcolor{cPass!70!black}{$\checkmark$} & \textcolor{cPass!70!black}{$\checkmark$} & \textcolor{cFail}{$\times$} & Best quality; slowest\\
\quad same, shuffled order & 99.8 & 0.2046 & $+.0198$ &  &  &  & \textcolor{cPass!70!black}{$\checkmark$} & \textcolor{cPass!70!black}{$\checkmark$} & \textcolor{cFail}{$\times$} & Keeps its lead\\
\addlinespace[2pt]\multicolumn{11}{@{}l}{\textit{Other decision models}}\\[1pt]
Winnow-12B, default + fallback & 100 & 0.2011 & $+.0168$ & 5,374$^b$ & 4.6$^b$ & -- & \textcolor{cPass!70!black}{($\checkmark$)} & \textcolor{cPass!70!black}{$\checkmark$} & \textcolor{cFail}{$\times$} & Too slow\\
Winnow-12B, truncated + fallback & 100 & 0.1875 & $+.0031$ & 2,892$^c$ & -- & -- & \textcolor{cPass!70!black}{($\checkmark$)} & \textcolor{cPass!70!black}{$\checkmark$} & \textcolor{black!55}{?} & Serving unverified\\
Kev-9B, truncated & 100 & 0.1213 & $-.0630$ & -- & -- & -- & \textcolor{cPass!70!black}{$\checkmark$} & \textcolor{cFail}{$\times$} & \textcolor{black!40}{--} & No gain over BM25\\
\bottomrule\end{tabular}
\tabnote{Rel.: $\ge$99\% of requests return a valid Top-5 from the model itself; ($\checkmark$) passes only if a BM25 fallback counts as completion. Qual.: beats the BM25 Top-5 and stays within 0.02 nDCG@5 of full-text Jina (lower 95\% bound above $-0.02$). Serv.: p95 at most $2\times$ full-text Jina's ($\checkmark$); ? = interval spans 2, not shown to pass; $\times$ = clearly above 2. Bold rows: PPLX policies passing all three. Latency of the budget-sweep rows: all 1,384 queries, two sessions with two passes each, one request at a time on an A100, next to a same-session full-text Jina anchor (two-stage 8,192: 120-query cohort); it includes failures and fallback time. Shuffled: mean over six candidate orders (measured at 8,192 tokens only). $^b$\,earlier shared session. $^c$\,p95 over completed requests only (failure times not recorded). --: not measured or not applicable. Tables~\ref{tab:sweep} and \ref{tab:original} give the full sweep and the default behavior.}
\end{table*}

\subsection{Does the result transfer across workloads?}\label{sec:overview}
\textbf{Two-stage PPLX ranks better than Jina on EnterpriseRAG-Bench and worse on MuSiQue, and neither result removes its latency disadvantage.} We reran the main policies on both benchmarks at the 8,192-token limit (Table~\ref{tab:replication}). Two-stage PPLX returns all annotated supporting candidates within its Top-5 more often than two-stage Jina on the enterprise benchmark, by $+0.102$ (interval $[+0.070,+0.135]$, Holm $p<.001$; $+0.132$ over the three other orders); Jina does better on MuSiQue, by $0.020$ (interval $[-0.031,-0.010]$, Holm $p=0.0003$; $-0.013$ over the other orders). Both PPLX policies stay \textbf{6.5 to 17 times} slower than full-text Jina at p95.

\begin{table}[t]
\centering
\caption{Do the replacement results hold across RAG workloads? Which model ranks better depends on the workload; the latency gap does not (controlled transfer check; FS@5; hybrid first-stage lists)}\label{tab:replication}
\scriptsize\setlength{\tabcolsep}{3.4pt}
\begin{tabular}{@{}l rr r r@{}}\toprule
 & Valid & & $\Delta$ vs full-text & p95\\
 Policy & (\%) & FS@5 & Jina & ratio\\
\midrule
\addlinespace[2pt]\multicolumn{5}{@{}l}{\textit{EnterpriseRAG-Bench (500 requests, 443 labeled; 74.5\% reachable)}}\\[1pt]
\quad First-stage order (hybrid) & 100 & 0.5576 & $-.0293$ & --\\
\quad Jina v3.5, full text & 100 & 0.5869 & ref. & 1.00\\
\quad Jina v3.5, truncated & 100 & 0.5327 & $-.0542$ & 0.52\\
\quad Jina v3.5, two-stage & 100 & 0.5688 & $-.0181$ & 1.2\\
\quad PPLX, default & 3.0 & 0.0226 & -- & --\\
\quad PPLX, truncated & 100 & 0.6524 & $+.0655$ & 6.5\\
\quad PPLX, two-stage & 100 & \textbf{0.6704} & $+.0835$ & 13.8\\
\quad \emph{PPLX $-$ Jina, both two-stage} & & & $\mathbf{+.1016}$ & \\
\addlinespace[2pt]\multicolumn{5}{@{}l}{\textit{MuSiQue (2,167 questions; 25.6\% reachable)}}\\[1pt]
\quad First-stage order (hybrid) & 100 & 0.0992 & $-.0978$ & --\\
\quad Jina v3.5, full text & 100 & 0.1970 & ref. & 1.00\\
\quad Jina v3.5, truncated & 100 & 0.1980 & $+.0009$ & 1.00\\
\quad Jina v3.5, two-stage & 100 & \textbf{0.2026} & $+.0055$ & 1.5\\
\quad PPLX, default & 100 & 0.1731 & -- & --\\
\quad PPLX, truncated & 100 & 0.1754 & $-.0217$ & 11.3\\
\quad PPLX, two-stage & 100 & 0.1823 & $-.0148$ & 17.3\\
\quad \emph{PPLX $-$ Jina, both two-stage} & & & $\mathbf{-.0203}$ & \\
\bottomrule\end{tabular}
\tabnote{Protocol frozen before outcomes. Bold FS@5: best policy on that benchmark. FS@5 is computed on labeled requests; a request is reachable when the Top-20 holds a complete support set (unreachable requests score 0 for every policy). Italic row: primary paired contrast on identical window text (Holm $p<.001$ and $p=.0003$). p95 ratio to full-text Jina in one dedicated session (120 requests $\times$ 3 per benchmark); truncated and two-stage PPLX exceed 2 on both. Lists come from a BM25 + dense fusion first stage, not BM25. Not comparable with BRIGHT nDCG@5.}
\end{table}

\textbf{A frozen test on a fresh benchmark repeats reliability and quality for both budgets, but latency only for 4,096 tokens.} On CQADupStack both PPLX settings returned a valid Top-5 for all 13,125 scored requests (exact lower bound 99.97\%; full-text Jina lost one to an out-of-memory error, which cannot move its mean by more than 0.0001, or 0.00005 on the forum mean). Pooled over queries, nDCG@5 is 0.2370 for the BM25 Top-5, 0.3198 for full-text Jina, 0.3290 for PPLX at 6,144 tokens and 0.3301 at 4,096; only 41.8\% of the Top-20 lists hold a judged positive, so judged-only scores are 2.4 times higher. Differences use the equal-weight mean over the 12 forums with forum-bootstrap intervals: the primary 6,144 setting beats the BM25 Top-5 by $+0.0900$ ($[+0.0784,+0.1022]$) and full-text Jina by $+0.0080$ ($[+0.0031,+0.0128]$), and 4,096 beats Jina by $+0.0092$ ($[+0.0039,+0.0143]$); both stay positive after correcting for two settings, and PPLX leads Jina in 10 and 11 of 12 forums. On a fixed 2,000-query subset, timed on one idle A100 in two serial, paired sessions, the p95 ratio to full-text Jina is \textbf{1.49 ($[1.29,1.69]$) at 4,096 tokens, a pass} (upper bound 1.72 at the two-setting 97.5\% level), and \textbf{2.19 ($[1.92,2.52]$) at 6,144, inconclusive} (97.5\%: $[1.88,2.56]$); Jina on shortened text is 0.30 times. The prespecified primary setting thus did not establish all three checks, while the secondary one passed all three, also at the corrected level.

\textbf{A corrected budget grid shows PPLX passing the quality check on Enterprise but not on MuSiQue.} We reran a budget grid with the frozen request builder, which reproduces the earlier 8,192-token inputs exactly (2,667 of 2,667 hashes) and keeps every request within $L-128$ tokens. Jina reads PPLX's strings; we score nDCG@5 on 443 labeled Enterprise and 2,167 MuSiQue questions, with failed requests at zero and a question-level bootstrap. A 2,048-token budget is infeasible on Enterprise (473 of 500 requests). On Enterprise, PPLX scores 0.590, 0.613 and 0.638 at 4,096, 6,144 and 8,192 tokens, against 0.332, 0.434 and 0.485 for same-text Jina and 0.538 for full-text Jina; it is above full-text Jina (lower bounds $+0.020$, $+0.045$, $+0.072$). On MuSiQue it scores 0.485, 0.482 and 0.487, against 0.536, 0.544 and 0.548 for same-text Jina and 0.547 for full-text Jina, so it trails full-text Jina by 0.060 to 0.065 (upper bounds below $-0.05$) and fails the quality check. We then timed the 2,667 pooled requests in two serial sessions each (matching anchors): PPLX's p95 is 1.10\,s at 4,096 and 1.7\,s at 6,144 tokens against 0.30\,s for full-text Jina, so it is 3.6 and 5.6 times slower (point estimates) and fails the latency check; Jina speeds up on these shorter candidate texts and PPLX does not. The BRIGHT passes carry over on quality for Enterprise only, and on latency nowhere.

These are controlled checks under a protocol fixed before the results were observed, not exact replications of BRIGHT. Both datasets had been examined before, the candidates come from a BM25 plus dense first stage, and the metric is FS@5 (the share of questions whose Top-5 holds a complete annotated support set), which is not comparable with nDCG@5 and is not answer correctness. A complete support set is reachable in the Top-20 for only 74.5\% of labeled Enterprise requests and 25.6\% of MuSiQue questions, which caps FS@5 for every policy. Truncating Jina's input costs $0.054$ (interval $[-0.079,-0.032]$) on Enterprise and nothing on MuSiQue, so truncation must be revalidated per workload. Which model ranks better depends on the workload; we do not isolate the cause.

\section{Limitations}\label{sec:lessons}
Our findings apply to the tested models, input policies and workloads. The 21-model screen follows heterogeneous historical protocols, and the controlled study centers on one outcome-selected pair whose sizes differ 45-fold (Section~\ref{sec:setup}). Relevance judgments are incomplete~\cite{buckley2004incomplete}, and the enterprise and multi-hop transfer sets were examined before and use different first-stage lists, so they are transfer checks. The frozen CQADupStack test adds independent evidence for reranking, but it is duplicate-question retrieval with sparse labels, not a measure of downstream answers, and its latency came from one GPU and 2,000 queries. We measure Top-K selection and serial latency on single A100s, not batched throughput or cost; our thresholds are illustrative, and the margin and fallback were chosen after earlier results. Other models, workloads or serving stacks may differ. The budget settings were chosen on one benchmark and confirmed only in part; statistical caveats on the sweep appear in Section~\ref{sec:sweep}.

\section{Conclusion}
\textbf{System-1-style decision models can perform competitive RAG reranking, but they are not inherently better replacements for dedicated rerankers.} Input handling decides whether a decision model is reliable, and a frozen evaluation on new data shows that one of two selected configurations meets all our replacement checks. Yet which model ranks better depends on the workload, and the gains can come with higher latency and far greater GPU memory use. Tuning the existing reranker offered the better trade-off. \textbf{For production RAG, tune the incumbent first, then test any alternative on representative candidate sets, and replace it only when the complete reranking policy meets the system's quality, reliability, latency and memory requirements.}

\section*{Acknowledgment}
\textit{AI-Assisted Preparation:} The authors used LLM assistants to write analysis scripts, draft figures and tables, and edit the text. The authors reviewed and validated the resulting analyses, figures, tables, and text and take full responsibility for the paper's content and findings.

\bibliographystyle{IEEEtran}
\bibliography{refs}

\end{document}